\pdfoutput=1
\documentclass[11pt]{article}

\usepackage[]{EMNLP2022}

\usepackage{times}
\usepackage{latexsym}

\usepackage[T1]{fontenc}
\usepackage[utf8]{inputenc}

\usepackage{microtype}

\usepackage{inconsolata}

\usepackage{graphicx} 
\usepackage{makecell}
\usepackage{amsmath}

\usepackage{multirow}
\usepackage{subfigure}

\usepackage{algorithm}
\usepackage{algorithmic}

\usepackage{afterpage}
\usepackage{lipsum}
\usepackage{float}
\usepackage{multicol} 

\title{Effective and Efficient Query-aware Snippet Extraction\\for Web Search}
\author{
  Jingwei Yi$^1$, Fangzhao Wu$^2$, Chuhan Wu$^3$, Xiaolong Huang$^4$, \\ \textbf{Binxing Jiao}$^4$, \textbf{Guangzhong Sun}$^1$, \textbf{Xing Xie}$^2$ \\
  $^1$University of Science and Technology of China
  $^2$Microsoft Research Asia \\
  $^3$Tsinghua University
  $^4$Microsoft STC Asia \\
  {\tt yjw1029@mail.ustc.edu.cn} {\tt \{wufangzhao,wuchuhan15\}@gmail.com} \\
  {\tt \{xiaolhu,binxjia,xingx\}@microsoft.com gzsun@ustc.edu.cn}
}

\begin{document}

\maketitle
\begin{abstract}

Query-aware webpage snippet extraction is widely used in search engines to help users better understand the content of the returned webpages before clicking.
Although important, it is very rarely studied.
In this paper, we propose an effective query-aware webpage snippet extraction method named DeepQSE, aiming to select a few sentences which can best summarize the webpage content in the context of input query.
DeepQSE first learns query-aware sentence representations for each sentence to capture the fine-grained relevance between query and sentence, and then learns document-aware query-sentence relevance representations for snippet extraction. 
Since the query and each sentence are jointly modeled in DeepQSE, its online inference may be slow.
Thus, we further propose an efficient version of DeepQSE, named Efficient-DeepQSE, which can significantly improve the inference speed of DeepQSE without affecting its performance.
The core idea of Efficient-DeepQSE is to decompose the query-aware snippet extraction task into two stages, i.e., a coarse-grained candidate sentence selection stage where sentence representations can be cached, and a fine-grained relevance modeling stage.
Experiments on two real-world datasets validate the effectiveness and efficiency of our methods.

\end{abstract}

\section{Introduction}

\begin{figure}[!t]
  \centering
  \includegraphics[width=0.48\textwidth]{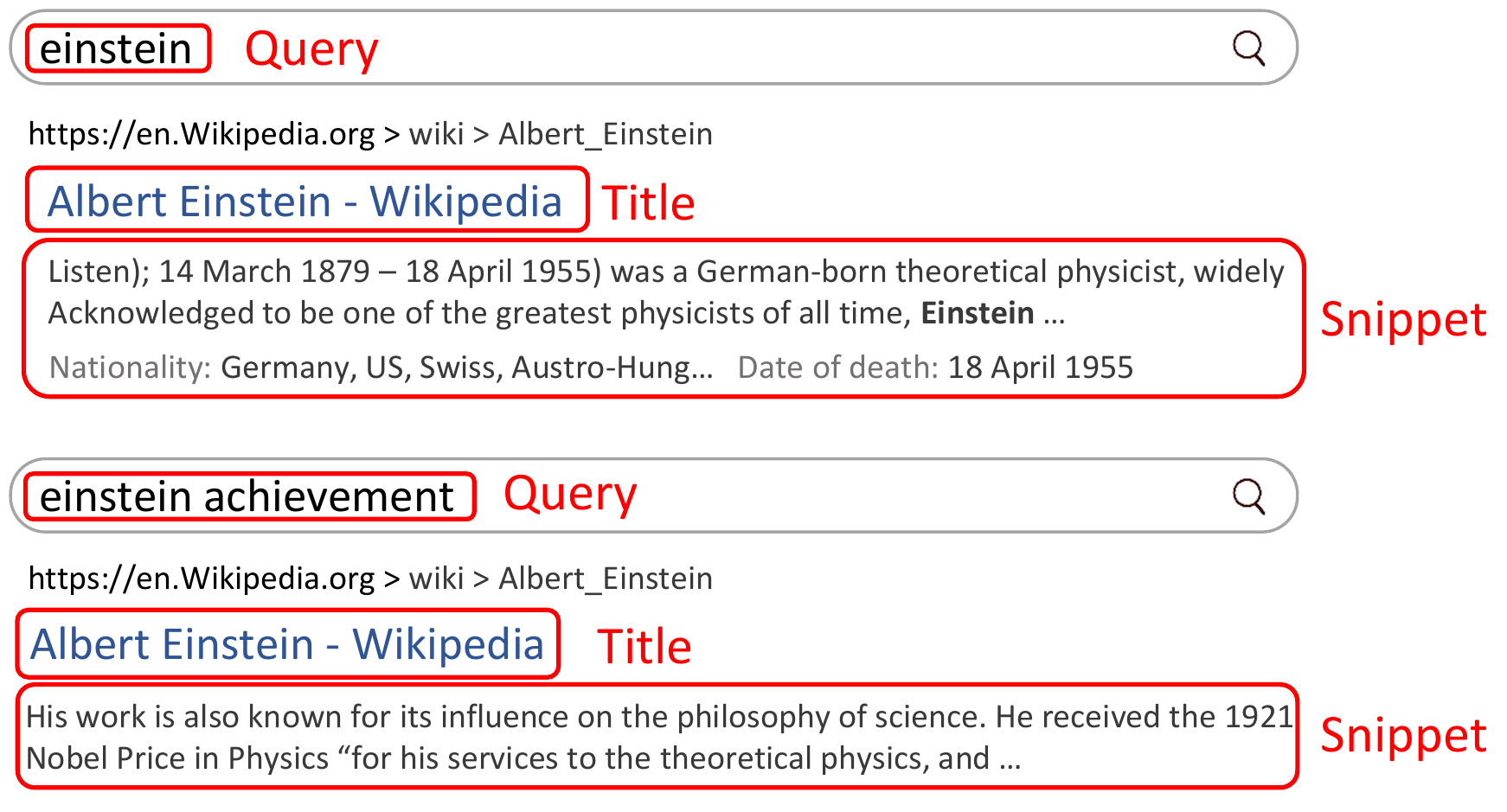}
  \caption{Examples of query-aware snippets in search engines.}
  \label{fig:snippet-example}
\end{figure}

Given an input search query, search engines such as Google~\footnote{https://www.google.com} and Bing~\footnote{https://www.bing.com}, not only return the URLs and the titles of the relevant webpages, but also show the query-aware snippets of these webpages, aiming to help users better understand the webpage content before clicking.
These webpage snippets are usually one or two sentences extracted from the webpage, which can not only summarize the key content of the webpage, but also be relevant to the input query.
Some examples are shown in Figure~\ref{fig:snippet-example}.
For the query `einstein' and the webpage of `Albert Einstein - Wikipedia', a good snippet is a brief introduction of Einstein's life. 
While for the query `einstein achievement', a good snippet would be sentences describing his influence on science.
In other words, the snippet is a summarization of the webpage in the context of input query. 

Although query-aware webpage snippet extraction is important and useful, it is very rarely studied.
Only a few works exist in this field, and most of them are based on simple word-level text matching method~\cite{10.1007/978-3-540-89704-0_34,zou2021pre}.
For example, ~\citet{10.1145/1277741.1277766} proposed to utilize the number of overlapping words between queries and sentences in webpages to extract snippets.
~\citet{tsegay2009document} proposed to select snippets through the summation of Kullback-Leibler divergence or TF-IDF weight of overlapping words between queries and sentences in webpages.
However, these methods rely on counting features of overlapping words, and cannot capture the deep semantic relation between query and webpage.

In this paper, we propose an effective query-aware webpage snippet extraction method for web search, named DeepQSE~\footnote{\url{https://github.com/yjw1029/DeepQSE}.}.
In DeepQSE, given an input query and a webpage with multiple sentences, we first learn query-aware sentence representations for each sentence to capture the fine-grained relevance between query, sentence and webpage title using a query-aware sentence encoder.
Then we model the query-sentence relevance in the context of the whole webpage using a document-aware relevance encoder.
Since the query and each webpage sentence are jointly modeled, the online inference speed of DeepQSE can be slow, while the search engines have extremely high requirements for low latency.
Thus, we further design an efficient version of DeepQSE named Efficient-DeepQSE,  aiming to significantly improve the inference speed of Deep-QSE and keep its performance as much as possible.
The key idea of Efficient-DeepQSE is to decompose the query-aware webpage snippet extraction task into two stages, i.e., coarse-grained candidate sentence selection and fine-grained relevance modeling.
The coarse-grained candidate sentence selection aims to select a moderate number of most potential sentences for snippet extraction using a bi-encoder where sentence representations can be cached for fast online serving.
The fine-grained relevance modeling stage aims to capture the deep semantic relevance between the query and the candidate sentences selected by the previous stage using query-aware cross-encoders.
We conducted many experiments on two real-world datasets, which verify the effectiveness and efficiency of our approach.
The contributions of this paper are as follows:
\begin{itemize}
    \setlength\itemsep{0.1em}
    \item We propose an effective query-aware webpage snippet extraction method for web search named DeepQSE, which can summarize the webpage content in the context of input query.
    \item We further propose Efficient-DeepQSE which can improve the inference speed of Deep-QSE with a minor performance drop.
    \item We conduct extensive experiments on two real-world datasets to verify the effectiveness and efficiency of our methods.
\end{itemize}

\section{Related Work}
% In this section, we first introduce some existing query-aware snippet extraction methods.
% Since the snippet extraction task can be partially modeled as computing the similarity between queries and sentences, we then introduce some text matching methods and compare the performance of our approach with them in Section~\ref{sec:exp}.

\subsection{Query-aware Snippet Extraction}
Query-aware snippet extraction is a widely-used technique to select snippets which can help users better understand the webpage content before clicking~\cite{10.1145/3366423.3380206,10.1145/1840784.1840813}.
Although important, only a few works have been proposed for query-aware snippet extraction based on word-level text matching method~\cite{zou2021pre,10.1145/1277741.1277766, 10.1007/978-3-540-89704-0_34,tsegay2009document}.
For example, ~\citet{10.1145/1277741.1277766} propose CTS, which selects snippets based on sentence positions and the number of overlapping words between queries and sentences.
~\citet{zou2021pre} propose QUITE, which computes importance scores for each word and sums the importance scores of overlapping words to select snippets.
These methods are mostly based on counting features of overlapping words and cannot capture deep semantic relations between query and webpage.
Recently, ~\citet{zhong-etal-2021-qmsum} propose QMSUM for meeting summarization, of which the locator can be used for snippet extraction.
The locator of QMSUM applies a fixed PLM and CNN to encode sentence and query, and a Transformer to model interactions between sentences.
QMSUM is a bi-encoder which fails to encoder the word-level interactions between query and sentences.
~\citet{zhao2021qbsum} propose QBSUM, which concatenates query and body, and applies multiple predictors to compute relevance scores.
The simple body-query concatenation in QBSUM may fluctuate the information of query and lead to some sentences being cut off due to the length limitation of PLM.
Recently, some works~\cite{10.1007/978-3-030-45442-5_22, 10.1145/3366423.3380206} use abstractive generation model to generate snippets for (query, document) pairs.
For example, ~\citet{10.1007/978-3-030-45442-5_22} uses the RNN network with copy mechanism to generate query-aware snippets.
% ~\citet{10.1145/3366423.3380206} uses a bidirectional LSTM network with an attention network to learn the (query, document) representation.
% Then, treating the query as the start word of the snippet, a pointer-generator network is used to complete the snippet in two directions, i.e. before or after the query.
However, abstractive methods need detailed parsing and digesting, which usually takes a considerable amount of time~\cite{10.1145/1321440.1321518}.
% Since the query-aware snippet extraction task for web search real-time process which is sensitive to computation efficiency, these methods are more suitable to perform general summarization which could be performed offline.
Therefore, these methods are not compared in this paper.

\subsection{Text Matching}
Text matching has been widely applied in many scenarios, such as information retrieval~\cite{10.1145/3132847.3132914} and clustering various articles for breaking news detection~\cite{Yang2002}.
Recently several text matching methods have been proposed.
Following ~\citet{Humeau2020Poly-encoders}, these methods can be divided into two groups, i.e., bi-encoders and cross-encoders.
Bi-encoders~\cite{palangi2014semantic,reimers-gurevych-2019-sentence, 10.5555/2969033.2969055} model the sentence-level interactions between queries and documents, in which the document representations can be cached for fast online serving.
For example, ~\citet{wan2016deep} propose C-DSSM, which computes query and document representations with convolutional networks.
Cross-encoders~\cite{guo2016deep,li2020parade, 10.1145/3219819.3219928} model the word-level interactions between queries and documents in a fine-grained manner.
For example, ~\citet{yilmaz2019cross} propose Birch, where (title, query) pairs are input into a pre-trained language model to compute matching scores.
Cross-encoders usually perform better than bi-encoders~\cite{urbanek-etal-2019-learning}, but have higher computation overhead since they cannot cache the document representations.
Since text matching methods can retrieve the most relevant sentence to query, we treat them as baseline methods and compare the performance with them in Section~\ref{sec:exp}.
However, the text matching methods only consider the similarity between queries and sentences, and ignore the contextual information of webpages, which might be sub-optimal.

\section{Methodology}
\begin{figure}[!t]
  \centering
  \includegraphics[width=0.40\textwidth]{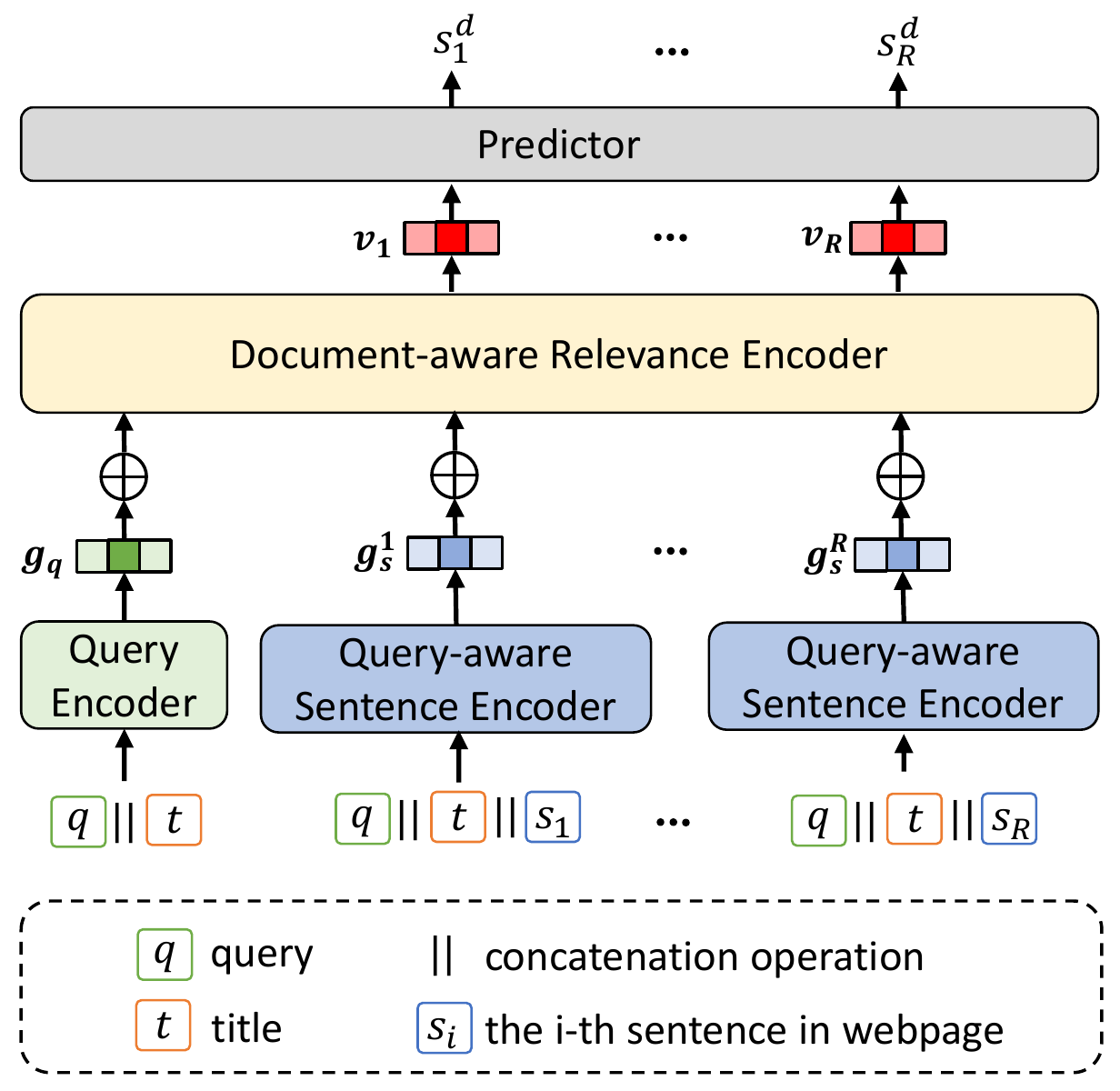}
  \caption{Architecture of DeepQSE.}
  \label{fig:DeepQSE}
\end{figure}
\begin{figure*}[!t]
  \centering
  \includegraphics[width=0.75\textwidth]{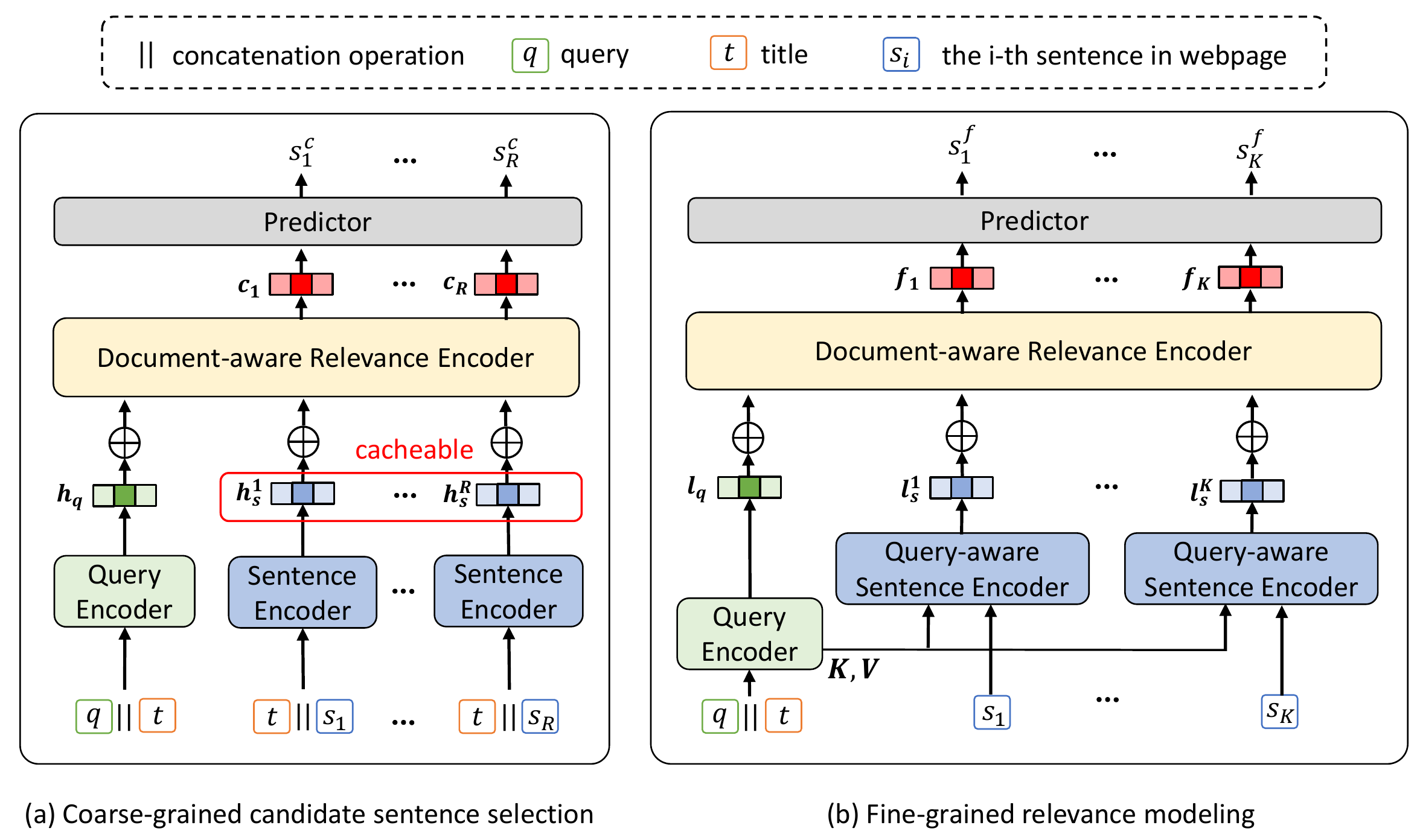}
  \caption{Framework of the two stages in Efficient-DeepQSE.}
  \label{fig:two_stage}
\end{figure*}
In this section, we give the problem formulation of query-aware snippet extraction.
Then we introduce our DeepQSE and Efficient-DeepQSE in detail.% In this section, we first introduce the model structure of our query-aware snippet extract model (DeepQSE). 
% Then, we introduce the efficient version of our model named Efficient-DeepQSE, which contains a coarse-grained candidate sentence selection stage and a fine-grained relevance modeling stage. 
% Finally, we introduce the training and serving procedure of our Efficient-DeepQSE.

\subsection{Problem Formulation}

When a user submits a request with query $q$, the search engine returns several webpages.
Given one of the webpage $d$ with title $t$, it contains several sentences $\{s_1, s_2, ... s_R\}$, where $R$ is the max number of sentences in a webpage.
The snippet extraction model aims to select several consecutive sentences $
\{s_k, ... s_{k+n}\}$ as the snippet that can summarize the webpage content in the context of input query.
Since the number of sentences $n$ is given by the pre-defined snippet length, the snippet extraction model needs to select the start sentence $s_k$.

\subsection{DeepQSE}
DeepQSE aims to select snippets which can best summarize the webpage content in the context of the input query.
The model structure of DeepQSE is shown in Figure~\ref{fig:DeepQSE}, which is composed of a query encoder, a query-aware sentence encoder and a document-aware relevance encoder.
The query encoder learns query representations, which is initialized from a pre-trained language model, such as BERT~\cite{devlin-etal-2019-bert} and XML-RoBERTa~\cite{chi-etal-2021-infoxlm}. 
Given the query $q$ and title $t$, the concatenation of them is input into the query encoder.
The final hidden state of the first token is the query representation $\textbf{g}_q$.
The query-aware sentence encoder models the word-level interactions between query, title and each sentence to compute query-aware sentence representations.
It is initialized from a pre-trained language model, of which the input is the concatenation of title, query and each sentence. 
The final hidden state of the first token is the sentence representation $\textbf{g}_s^i$.
The document-aware relevance encoder aims to model the sentence-level interactions between the query and sentences in the context of the whole webpage, which is composed of several Transformer blocks~\cite{NIPS2017_3f5ee243}.
We concatenate the query representation and sentence representations, add position embeddings and input them into the document-aware relevance encoder.
The final hidden states are used as document-aware query-sentence relevance representations $\textbf{v}_i$, which are then used to compute the selection score $s_i^d$.
% However, the DeepQSE is a cross-encoder with high computation overhead.
% To reduce the computation cost of DeepQSE while keeping its performance, we further design its efficient version, i.e, Efficient-DeepQSE.

\subsection{Efficient-DeepQSE}
In DeepQSE, the query and each sentence are jointly modeled, which may have slow computation speed for online serving.
In order to reduce the computation overhead, we further design an efficient version of DeepQSE named Efficient-DeepQSE, which is shown in Figure~\ref{fig:two_stage}.
We decompose the query-aware snippet extraction into two stages, i.e., coarse-grained candidate sentence selection and fine-grained relevance modeling.

% \begin{figure*}[!t]
% \centering 
% \resizebox{0.80\linewidth}{!}{
%     \centering 
%     % \subfigure[Coarse-grained selector.]{\label{fig:QaSE} 
%     % \begin{minipage}[t]{0.333\textwidth}
%     %     \centering
%     %     \includegraphics[width=\textwidth]{figure/QaSE.pdf} 
%     % \end{minipage}
%     % } 
%     \subfigure[Coarse-grained candidate selector.]{\label{fig:coarse-selector} 
%     \begin{minipage}[t]{0.35\textwidth}
%         \centering
%         \includegraphics[width=\textwidth]{figure/coarse-selector.pdf} 
%     \end{minipage}
%     } \hspace{0.02\linewidth}
%     \subfigure[Fine-grained relevance model]{\label{fig:fine-selector}
%     \begin{minipage}[t]{0.343\textwidth}
%         \centering
%         \includegraphics[width=\textwidth]{figure/fine-selector.pdf}
%     \end{minipage}
%     }
% }
% \caption{Framework of the two stages in Efficient-DeepQSE.} 
% \label{fig:two_stage}
% \end{figure*}

\subsubsection{Coarse-grained Sentence Selection}
The coarse-grained candidate sentence selection aims to select $K$ candidate sentences and parse them to the fine-grained relevance model for final snippet extraction.
It separates the modeling of candidate sentences and queries, which enables caching sentence representations for fast online serving. 
The model structure of the coarse-grained candidate selector is shown in Figure~\ref{fig:two_stage}(a), which contains three core modules, i.e., a query encoder, a sentence encoder and a document-aware relevance encoder.
The query encoder and sentence encoder aim to learn the query and sentence representations respectively, which are initialized from a pre-trained language model.
We input the concatenation of query and title into the query encoder, and use the final hidden states of the first token as the query representation $\textbf{h}_q$.
The concatenation of title and each sentence is input into the sentence encoder, and the final hidden states of the first token are treated as the sentence representation $\textbf{h}_s^i$.
% The input of sentence encoder is the concatenation of title and each sentence, and the final hidden state of the first token is the sentence representation $\textbf{h}_s^i$.
The document-aware sentence relevance encoder aims to model query-sentence relevance in the context of the whole webpage, which is composed of several Transformer blocks.
We concatenate the query representation and sentence representations, add position embeddings and input them into the document-aware sentence relevance encoder.
The final hidden states are the document-aware query-sentence relevance representations $\textbf{c}_i$, which are further used to predict selection scores $s_i^c$ through an MLP network.

\subsubsection{Fine-grained Relevance Modeling}
\begin{figure*}[!t]
  \centering
  \includegraphics[width=0.80\textwidth]{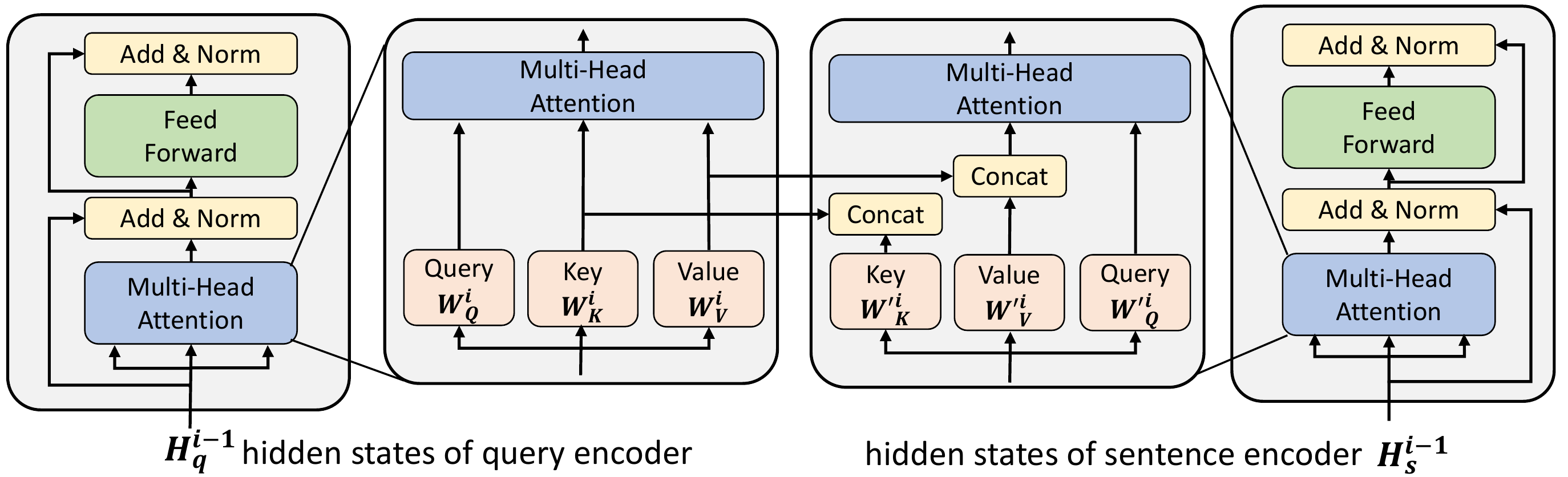}
  \caption{Architecture of the Cross Transformer model.}
  \label{fig:cross-tsf}
\end{figure*}

The fine-grained relevance modeling aims to capture the deep semantic relevance between query and the candidate sentences parsed from the coarse-grained sentence selection stage.
It is composed of a query encoder, a query-aware sentence encoder and a document-aware relevance encoder, of which the model structure is shown in Figure~\ref{fig:two_stage}(b).
% The query encoder is a Transformer-based pre-trained language model (PLM).
% Given query $q$ and title $t$, we concatenate them into a word sequence $[[CLS], w_1^q,...w_N^q, [SEP], w_1^s, w_L^s]$.
% Then we compute their word embeddings, add position embeddings~\cite{} and input them into several transformer blocks.
% The final hidden state of the $[CLS]$ token is the query representation $\textbf{h}_q$.
A naive implementation is directly using the same architecture of DeepQSE.
However, in DeepQSE, the query and title are concatenated with different $R$ sentences and their word representations are repetitively computed for $R$ times.
We assume the word representations of query in the query-aware sentence encoder have little help for query-aware snippet selection, which is validated in Section~\ref{ref:efficient}.
% In order to model the word level interaction between title, query and sentence, we can concatenate them into a sequence, and input it into a pre-trained language model.
% The sentence representation can be the final hidden state of the $[CLS]$ token.
% However, we notice that the word representations of the query and title are repeatedly computed for the $R$ sentences in a document.
Therefore, we design the Cross Transformer, where the word representations of the query and title are only computed once in the query encoder and parsed into the query-aware sentence encoder.
The architecture of a Cross Transformer block is shown in Figure~\ref{fig:cross-tsf}.

The query encoder aims to learn query representations, which is initialized from a Transformer-based pre-trained language model.
We input the concatenation of query and title into the query encoder and use the final hidden state of the first token as the query representation $\textbf{l}_q$.
Meanwhile, the query encoder outputs the key and value of every multi-head self attention network to the query-aware sentence encoder.
Given the previous hidden state $\textbf{H}^{i-1}_{q}$, the key and value of the $i$-th multi-head self attention network are computed as follows:
\begin{equation}
    \textbf{K}_i^q = \textbf{W}_K^i\textbf{H}^{i-1}_{q}, \textbf{V}_i^q=\textbf{W}_V^i\textbf{H}^{i-1}_{q},
\end{equation}
where $\textbf{W}_K^i$ and $\textbf{W}_V^i$ are trainable parameters.
% The key and value are then input into the multi-head self attention network and the feed-forward network to compute $\textbf{H}^{i}_{q}$.
% We use the final hidden state of the first token as the query representation $\textbf{l}_q$.
% The key and value are also parsed to sentence encoder to model the word-level interaction between the query, title and sentences.

The query-aware sentence encoder aims to model the fine-grained interactions between query and each sentence. 
It contains an embedding layer and several Cross Transformer blocks, which are initialized from a pre-trained language model.
Given a sentence $s_i$, we first compute its initial hidden state $\textbf{H}_s^{0}$ through the embedding layer.
% We input the sentence tokens, the key and value parsed from the query encoder into the sentence encoder.
The $i$-th Cross Transformer block contains a multi-head attention network and a feed-forward network.
In order to compute query-aware sentence hidden states, we modify the key (or value) of multi-head attention network as the concatenation of key (or value) from query encoder and the transformed hidden state of the sentence.
% The same operation is applies on the value of multi-head attention network as well.
% In order to compute query-aware sentence representations, we modify the key and query of multi-head attention network to let each word encode relevant information in both query, title and sentence.
Given the hidden state $\textbf{H}_s^{i-1}$ of the previous Cross Transformer block, the query, key and value of the $i$-th multi-head attention network are formulated as follows:
\begin{equation}
    \begin{aligned}
        \textbf{Q}_i^s &= \textbf{W}_Q^{i'}\textbf{H}_s^{i-1},\\
        \textbf{K}_i^s &= \textbf{K}_i^q||\textbf{W}_K^{i'}\textbf{H}_s^{i-1},\\
        \textbf{V}_i^s &= \textbf{V}_i^q||\textbf{W}_V^{i'}\textbf{H}_s^{i-1},
    \end{aligned}
\end{equation}
where $\textbf{W}_Q^{i'}$, $\textbf{W}_K^{i'}$ and $\textbf{W}_V^{i'}$ are trainable parameters, $||$ is the concatenation operator.
The query, key and value are then input into the multi-head attention network and feed-forward network to compute $\textbf{H}_s^i$.
% In this way, the multi-attention attention network computes query-aware word representations for each sentence.
We use the final hidden state of the first token as the query-aware sentence representation $\textbf{l}_s^i$.

The document-aware relevance encoder aims to model query-sentence relevance in the context of the whole webpage, which contains several Transformer blocks.
We concatenate the query representations and $K$ candidate sentence representations, add position embeddings on them and input them into several transformer blocks.
The document-aware query-sentence relevance representations $\textbf{f}_i$ are the final hidden states, which are then fed into an MLP to predict selection scores $s_i^f$.

\subsubsection{Model Training and Serving}

% In this subsection, we introduce the model training and serving processes. 
For model training, we use cross-entropy loss to train the DeepQSE, which is computed as follows:
\begin{equation}
    \mathcal{L} = -\sum_{i=1}^{R}y_i\times log(\frac{exp(s_i^d)}{\sum_{k=1}^{R}exp(s_k^d)}),
\label{eq:loss}
\end{equation}
where $y_i \in \{0, 1\}$ indicates whether the $i$-th sentence is the start sentence of the snippet.
For Efficient-DeepQSE, we also use the cross-entropy loss to train the coarse-grained candidate sentence selector and the fine-grained relevance model respectively, which is formulated as follows:
\begin{equation}
    \begin{aligned}
        \mathcal{L}_c &= -\sum_{i=1}^{R}y_i\times log(\frac{exp(s_i^c)}{\sum_{k=1}^{R}exp(s_k^c)}), \\
        \mathcal{L}_f &= -\sum_{i=1}^{R}y_i\times log(\frac{exp(s_i^f)}{\sum_{k=1}^{R}exp(s_k^f)}).
    \end{aligned}
\label{eq:loss_c}
\end{equation}
For model serving, when a user submits a request with query $q$, the search engine returns several webpages.
For one of the webpages $d$ with title $t$, DeepQSE directly computes selection scores for all sentences $\{s_1^d, ... s_R^d\}$.
The sentence with the maximum selection score is selected.
% The model serving algorithm of Efficient-DeepQSE is shown in Algorithm~\ref{alg:serve}.
For Efficient-DeepQSE, the server needs to offline compute the sentence representations of coarse-grained candidate sentence selector for every webpage.
For the webpage $d$, with its sentence representations of the coarse-grained candidate sentence selector $[\textbf{h}_s^1, ...\textbf{h}_s^R]$, we first compute the query representation of the coarse-grained candidate sentence selector $\textbf{h}_q$ and the coarse-grained selection scores $\{s_1^c, ... s_R^c\}$.
Then we feed top-K candidate sentences into the fine-grained relevance model to compute fine-grained selection scores $\{s_1^f, ... s_K^f\}$.
The sentence with the maximum score is the start sentence of the snippet $s_k$.

\begin{table}[!t]
\centering
% \resizebox{\linewidth}{!}{
\scalebox{0.80}{
\begin{tabular}{c|cc}
\Xhline{1.5pt}
                % \multicolumn{3}{c}{Snippet extraction}  \\ \hline
Snippet extraction dataset  & \textit{English}             & \textit{Multi-lingual}     \\ \hline
\#sample        &  545,240            &     348,673               \\
\#query         &  420,816            &     291,559               \\
\#document      &  330,545            &     240,005               \\ \hline
% \multicolumn{3}{c}{Manually labeled} \\ \hline
Manually labeled dataset & \textit{English}             & \textit{Multi-lingual}     \\ \hline
\#sample        &  19,331            & 25,851                  \\
\#query         &  14,522            & 17,935                  \\
\#document      &  16,995            & 22,726                  

\\ \Xhline{1.5pt}
\end{tabular}
}
\caption{Statistics of datasets.}
\label{tab:stat}
\end{table}
\begin{table*}[!t]
\centering
\scalebox{0.85}{
\begin{tabular}{c|ccc|ccc}
\Xhline{1.5pt}
\multirow{2}{*}{Methods} & \multicolumn{3}{c|}{\textit{English}}    & \multicolumn{3}{c}{\textit{Multi-lingual}} \\ \cline{2-7} 
                         & P@1 & P@3 & P@5  & P@1  & P@3  & P@5  \\ \hline
CTS                      & 39.65±0.00 & 64.57±0.00 &  88.15±0.00  & 34.64±0.00  & 59.83±0.00  & 71.16±0.00   \\
QUITE                    & 39.49±0.00 & 63.69±0.00 &  74.78±0.00  & 33.71±0.00  & 57.24±0.00  & 68.96±0.00   \\ \hline
QMSUM                    & 54.22±0.27 & 74.61±0.20 &  82.17±0.24  & 46.26±0.27  & 67.19±0.18  & 75.87±0.17   \\
QBSUM                    & 60.69±0.51 & 77.18±0.26 &  81.12±0.38  & 59.61±0.77  & 71.80±0.99  & 74.84±0.67   \\ \hline
BM25                     & 33.91±0.00 & 60.50±0.00 &  73.15±0.00  & 27.75±0.00  & 51.99±0.00  & 65.48±0.00   \\
DSSM                     & 40.24±0.50 & 59.98±0.59 &  69.12±0.54  & 36.88±0.30  & 54.33±0.34  & 63.29±0.29   \\
C-DSSM                   & 55.46±0.31 & 72.73±0.23 &  79.39±0.26  & 49.49±0.41  & 66.32±0.37  & 73.80±0.30   \\
MatchPyramid             & 55.49±0.40 & 78.04±0.40 &  85.78±0.26  & 50.74±0.35  & 73.58±0.30  & 82.05±0.39   \\ \hline
Poly-Encoder             & 64.45±0.11 & 82.19±0.08 &  88.02±0.11  & 64.41±0.13  & 82.14±0.10  & 88.00±0.07   \\
Birch                    & 72.45±0.08 & 88.24±0.10 &  92.73±0.08  & 72.62±0.11  & 88.44±0.14  & 92.88±0.06   \\ 
PARADE                   & 73.19±0.16 & 88.65±0.12 &  92.99±0.14  & 72.94±0.20  & 88.47±0.13  & 92.89±0.13   \\  \hline
DeepQSE*                 & \textbf{77.05}±0.29 & \textbf{92.43}±0.33 &  95.94±0.14  & \textbf{77.23}±0.18  & \textbf{93.30}±0.09  & \textbf{96.77}±0.08   \\
Efficient-DeepQSE*        & \textbf{77.03}±0.27 & \textbf{91.98}±0.19 &  95.34±0.13  & \textbf{75.13}±0.27  & \textbf{91.40}±0.22  & \textbf{95.15}±0.14   \\ \Xhline{1.5pt}
\end{tabular}
}
\caption{Performance of different methods on query-aware webpage snippet extraction.}
\label{tab:experiment}
\end{table*}
\section{Experiments}
\label{sec:exp}
\subsection{Dataset and Experimental Settings}
Since there is no off-the-shelf dataset for query-aware snippet extraction, we first manually labeled two small \textit{English} and \textit{Multi-lingual} datasets, of which the task is to select the more proper snippet given a pair of candidate sentences.
The \textit{Multi-lingual} dataset includes 10 languages, i.e., German, French, Spanish, Italian, Japanese, Korean, Portuguese, Russian and Chinese.
Then we semi-automatically build two large \textit{English} and \textit{Multi-lingual} snippet extraction datasets with part of the manually-labeled datasets, of which task is to select the snippet from the sentences in the body.
Due to the space limitation, the detailed dataset construction steps are described in Appendix~\ref{sec:appendix-data}.
10\% samples of the snippet extraction dataset are randomly sampled for testing, and the rest for training.
We randomly sample 10\% samples of the training dataset for validation.
We also the rest manually labeled dataset as another test dataset, which is not used to construct the large snippet extraction dataset.
% It is noted that we keep part of the manually labeled dataset as test dataset, which is not used to construct the enlarged dataset.
% Since the whole test dataset is manually labeled, it can be treated as a golden ground truth.
The detailed statistics of the datasets are shown in Table~\ref{tab:stat}.
We use precision@k (k=1,3,5) as the evaluation metrics for performance on the snippet extraction test dataset, accuracy (ACC) as the evaluation metric for performance on the manually labeled test dataset, floating-point operations (FLOPs) and million seconds (ms) as the evaluation metrics for efficiency.

\subsection{Experimental Settings}
In our experiments, we apply BERT-base~\cite{devlin-etal-2019-bert} for \textit{English} dataset and a distilled XML-RoBERTa~\cite{chi-etal-2021-infoxlm} for \textit{Multi-Lingual} dataset to initialize the pre-trained language model.
We use Adam~\cite{kingma2014adam} to optimize model training for both DeepQSE and Efficient-DeepQSE.
We set the learning rate as 0.0001 and batch size as 64.
The maximum query length is 16.
The maximum title length is 32.
The maximum sentence length is 64.
The maximum number of sentences $R$ in a body is 160.
The number of candidate sentences selected by the coarse-grained sentence selector $K$ is 20.
All hyper-parameters are selected according to the performance on the validation dataset.
We repeat each experiment 5 times and report the average results and the standard deviations.

\subsection{Performance Comparison}
\label{sec:perform-comp}
We compare our method with multiple baselines, including conventional snippet extraction methods:
(1) CTS~\cite{10.1145/1277741.1277766}, extracting snippets based on the number of overlapping words between queries and sentences;
(2) QUITE~\cite{zou2021pre}, selecting snippets with the summation of importance scores of overlapping words between queries and sentences;
PLM-empowered snippet extraction methods:
(3) QMSUM~\cite{zhong-etal-2021-qmsum}, the locator of QMSUM for meeting summarization which applies a fixed PLM and CNN to encode sentence and query, and a Transformer to model interactions between sentences.
(4) QBSUM~\cite{zhao2021qbsum}, input the concatenation of query and body into a PLM, and apply multiple predictors to compute relevance scores;
some text matching methods:
(5) BM25~\cite{10.1561/1500000019}, applying the BM25 algorithm to compute similarity scores;
(6) DSSM~\cite{huang2013learning}, a deep structured semantic matching method;
(7) C-DSSM~\cite{wan2016deep}, a deep semantic matching structure with convolution network;
(8) MatchPyramid~\cite{pang2016text}, applying 2D convolution and max-pooling network on the similarity matrix of query and sentence;
several PLM-empowered text matching methods:
(9) Poly-Encoder~\cite{Humeau2020Poly-encoders}, which adds a final attention mechanism to model the interactions between the cacheable multiple sentence representations and the query representation.
(10) Birch~\cite{yilmaz2019cross}, inputting the concatenation of queries and sentences into BERT for document retrieval;
(11) PARADE~\cite{li2020parade}, using a PLM to model similarity between sentences and queries, and an aggregator to model interactions between candidate sentences.

\begin{table}[!t]
\centering
\scalebox{0.85}{
\begin{tabular}{c|cc}
\Xhline{1.5pt}
\multirow{2}{*}{Methods} & \multicolumn{2}{c}{ACC}                                        \\ \cline{2-3} 
                         & \multicolumn{1}{c|}{\textit{English}}             & \textit{Multi-lingual}       \\ \hline
CTS                      & \multicolumn{1}{c|}{29.20±0.00}          & 29.84±0.00          \\
QUITE                    & \multicolumn{1}{c|}{24.82±0.00}          & 25.73±0.00          \\ \hline
QMSUM                    & \multicolumn{1}{c|}{38.05±0.16}          & 39.61±0.21          \\
QBSUM                    & \multicolumn{1}{c|}{21.73±0.86}          & 26.26±0.65          \\ \hline
BM25                     & \multicolumn{1}{c|}{33.10±0.00}          & 33.11±0.00          \\
DSSM                     & \multicolumn{1}{c|}{36.67±0.46}          & 37.16±0.18          \\
C-DSSM                   & \multicolumn{1}{c|}{36.97±0.26}          & 37.77±0.21          \\
MatchPyramid             & \multicolumn{1}{c|}{35.54±0.16}          & 38.28±0.16          \\ \hline
Poly-Encoder             & \multicolumn{1}{c|}{37.51±0.23}          & 39.82±0.22          \\
Birch                    & \multicolumn{1}{c|}{38.17±0.28}          & 39.83±0.21          \\
PARADE                   & \multicolumn{1}{c|}{36.68±0.17}          & 38.59±0.13          \\ \hline
DeepQSE*                 & \multicolumn{1}{c|}{\textbf{40.10}±0.58} & \textbf{41.99}±0.20 \\
Efficient-DeepQSE*       & \multicolumn{1}{c|}{\textbf{40.57}±0.34} & \textbf{41.99}±0.26 \\ \Xhline{1.5pt}
\end{tabular}
}
\caption{Performance of different methods on manually labeled datasets.}
\label{tab:exp-labeled}
\end{table}

The performance of all methods on snippet extraction test datasets is shown in Table~\ref{tab:experiment}.
The performance of the methods on manually labeled test datasets is shown in Table~\ref{tab:exp-labeled}.
CTS, QUITE and BM25 are deterministic methods, of which standard deviations are zero.
We have several observations from Table~\ref{tab:experiment}.
First, our DeepQSE and Efficient-DeepQSE outperform conventional snippet extraction methods (CTS and QUITE).
This is because these methods are based on the counting features of overlapping words between queries and sentences.
Compared with our methods which utilize PLMs, they cannot capture the deep semantic relation between query and sentences.
Second, our methods outperform PLM-based snippet extraction methods (QMSUM and QBSUM).
This is because the simple body-query concatenation in QBSUM may fluctuate the information of the short query.
Due to the length limitation of PLM some candidate sentences may be cut off.
QMSUM is a bi-encoder which fails to encoder the word-level interactions between the query and sentences.
Third, compared with several text matching methods (BM25, DSSM, C-DSSM, MatchPyramid, Poly-Encoder, Birch, QBSUM), our methods have better performance.
This is because in our methods we apply webpage title and document-aware relevance encoder to select snippets in the context of the whole webpage, which can choose sentences better summarizing the webpage in the context of input query.
% Meanwhile, we add the document-aware relevance encoder which can help model both the sentence-level relevance between queries, titles and sentences and compute contextual sentence representations.
Forth, PLM-based snippet extraction methods outperform conventional snippet extraction methods, and PLM-based text-matching methods outperform shallow-model-based text-matching methods.
This is because the pre-trained language model can help better understand the semantic information in queries, titles and sentences.
Finally, cross-encoder-based text matching methods outperform bi-encoder-based text matching methods.
For example, MatchPyramid outperforms CDSSM and DSSM, and Birch, PARADE and DeepQSE outperform Poly-Encoder and QBSUM. 
This is because bi-encoders only model sentence-level similarity between queries and sentences, but cross-encoders can model word-level similarity between queries and sentences in a fine-grained manner.
However, bi-encoders can cache sentence representations, which have faster online serving speed than cross-encoders.

% \subsection{Efficiency Comparison (RQ2)}

% \begin{table}[!t]
% \centering
% \scalebox{0.85}{
% \begin{tabular}{c|c|c}
% \Xhline{1.5pt}
% \multirow{2}{*}{Methods} & English & Multi-Lingual \\ \cline{2-3} 
%                          & GFLOPs  & GFLOPs        \\ \hline
% DSSM                     & $2.71 \times 10^{-4}$   &  $2.72 \times 10^{-4}$       \\
% CDSSM                    & $6.58 \times 10^{-3}$   &  $7.66 \times 10^{-3}$       \\
% MatchPyramid             & 0.10                    &  0.10         \\ \hline
% Birch                    & 130.90                  &  40.67        \\
% PARADA                   & 131.38                  &  40.81        \\ \hline
% Efficient-SG             & 27.46                   &  8.12        \\ \Xhline{1.5pt}
% \end{tabular}
% }
% \caption{Efficiency of different methods.}
% \label{tab:efficiency}
% \end{table}

% We further compare the efficiency of our methods with PLM-based baseline methods.
% The computation overheads of different methods are shown in Table~\ref{tab:efficiency}.
% From Table~\ref{tab:efficiency} we can observed that the computation overhead of our Efficient-SE is lower than others.
% This is because we apply two-stage model, where the coarse-grained selector can efficiently select $K$ candidate sentences and parse them into the fine-grained selector for accurate decision.
% Moreover, we design the Cross Transformer, which can avoid the repetitive computation of queries for different sentences.
\begin{table}[!t]
\scalebox{0.75}{
\begin{tabular}{c|cc|cc}
\Xhline{1.5pt}
\multirow{2}{*}{Methods} & \multicolumn{2}{c|}{\textit{English}} & \multicolumn{2}{c}{\textit{Multi-lingual}} \\ \cline{2-5} 
                         & FLOPs               & ms              & FLOPS                  & ms                \\ \hline
CTS                      & -                   &  1.10           & -                      &  1.51             \\
QUITE                    & -                   &  0.13           & -                      &  0.10             \\ \hline
QBSUM                    & 45.75G              &  7.97           &    11.40G              &  4.02             \\
QMSUM                    & 1.74G               &  0.41           &    0.43G               &  0.17             \\ \hline
BM25                     & -                   &  0.80           & -                      &  0.95             \\
DSSM                     & 0.30M               &  0.18           &    0.30M               &  0.09             \\
C-DSSM                   & 17.42M              &  0.17           &    17.42M              &  0.09             \\
MatchPyramid             & 0.28G               &  0.37           &    0.28G               &  0.38             \\ \hline
Poly-Encoder             & 1.55G               &  0.19           &    0.43G               &  0.08             \\
Birch                    & 1087.44G            &  21.72          &    271.86G             &  10.51            \\
PARADE                   & 1088.71G            &  22.35          &    272.17G             &  11.16            \\ \hline
DeepQSE*                 & 1540.08G            &  31.91          &    271.86G             &  16.70            \\
Efficient-DeepQSE*       & 132.45G             &  3.09           &    33.44G              &  1.67             \\
\Xhline{1.5pt}
\end{tabular}
}
\caption{Efficiency of different methods.}
\label{tab:exp-eff}
\end{table}
\subsection{Efficiency Comparison}

In this subsection, we compare the efficiency of DeepQSE and Efficient-DeepQSE with baseline methods.
The results are summarized in Table~\ref{tab:exp-eff}.
Since CTS, QUITE and BM25 are not based on matrix multiplication and addition, we do not give their FLOPs results.
We have several observations from Table~\ref{tab:exp-eff}.
First, conventional snippet extraction methods (CTS and QUITE) have relatively low computation costs.
This is because they are based on simple hand-crafted features, which can be calculated quickly.
Second, cross-encoder-based methods are more time-consuming than bi-encoder-based methods.
For example, DSSM and CDSSM are more efficient than MatchPyramid, and Poly-Encoder and QMSUM are more efficient than Birch, PARADE, QBSUM and DeepQSE.
This is because the sentence representations in bi-encoder-based methods can be cached for quick inference.
Third, PLM-based methods (Poly-Encoder, Birch, PARADE, QMSUM, QBSUM, DeepQSE and Efficient-DeepQSE) have higher computation costs than other methods.
This is because pre-trained language models have large size of parameters, of which the computation cost is high~\cite{DBLP:journals/corr/abs-1910-01108,beltagy2020longformer,jiao-etal-2020-tinybert,sun2020mobilebert}.
Finally, considering both efficiency and the previous performance analysis in Section~\ref{sec:perform-comp}, our Efficient-DeepQSE achieves a great trade-off between performance and efficiency.
This is because our Efficient-DeepQSE applies two-stage model, in which the coarse-grained selector can quickly select candidates for the fine-grained relevance encoder.
In addition, we design the Cross Transformer which avoids repetitively computing contextual word representations of the same query for different candidate sentences.
Therefore, our Efficient-DeepQSE has a lower computation cost while keeping its performance.

\subsection{Efficiency Analysis}
\begin{figure}[!t]
  \centering
  \includegraphics[width=0.40\textwidth]{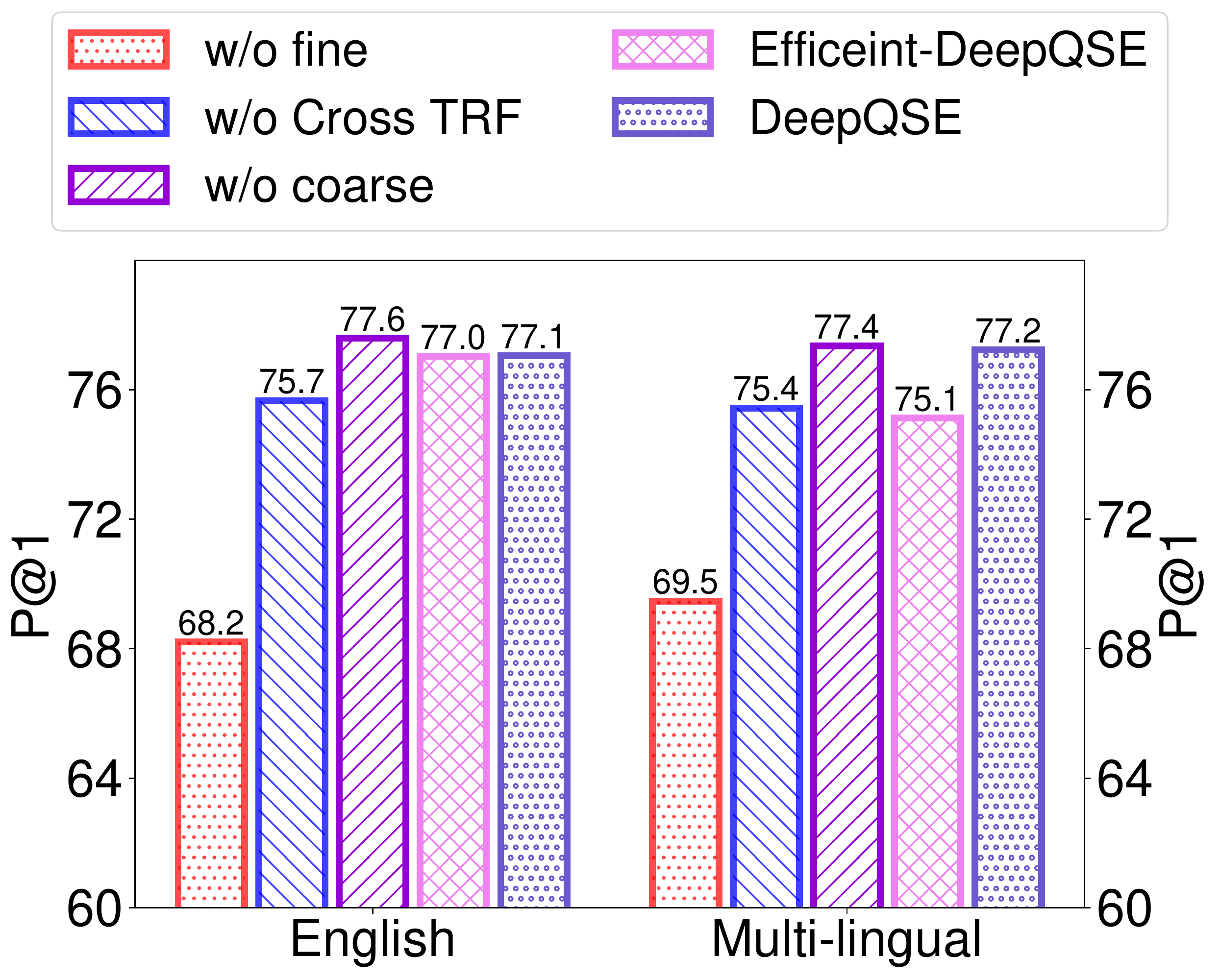}
  \caption{The impact of two-stage model and Cross Transformer on accuracy.}
  \label{fig:two-step-perform}
\end{figure}

\begin{figure}[!t]
  \centering
  \includegraphics[width=0.40\textwidth]{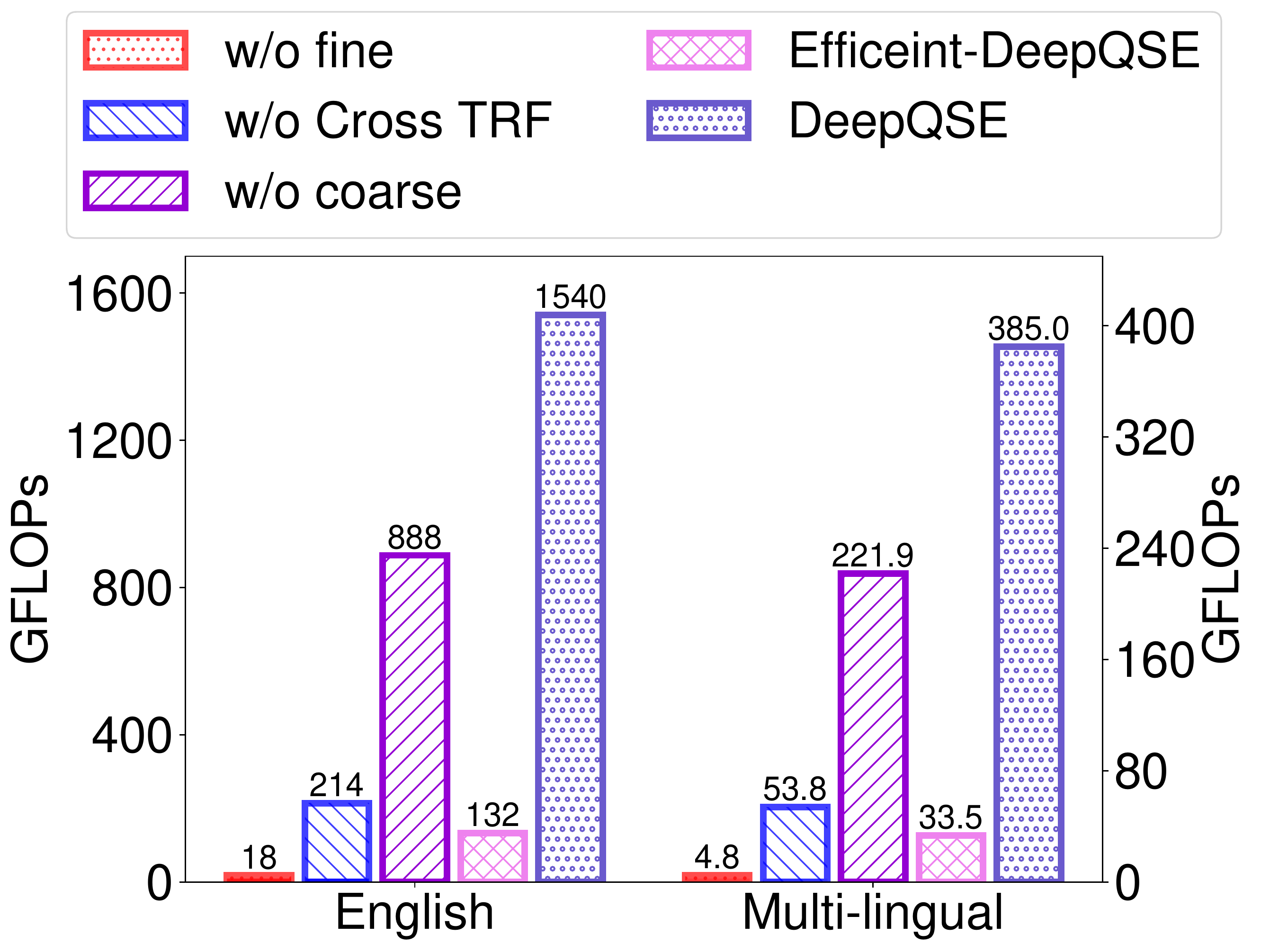}
  \caption{The impact of two-stage model and Cross Transformer on efficiency (GFLOPs).}
  \label{fig:two-step-efficiency}
\end{figure}

\begin{figure}[!t]
  \centering
  \includegraphics[width=0.40\textwidth]{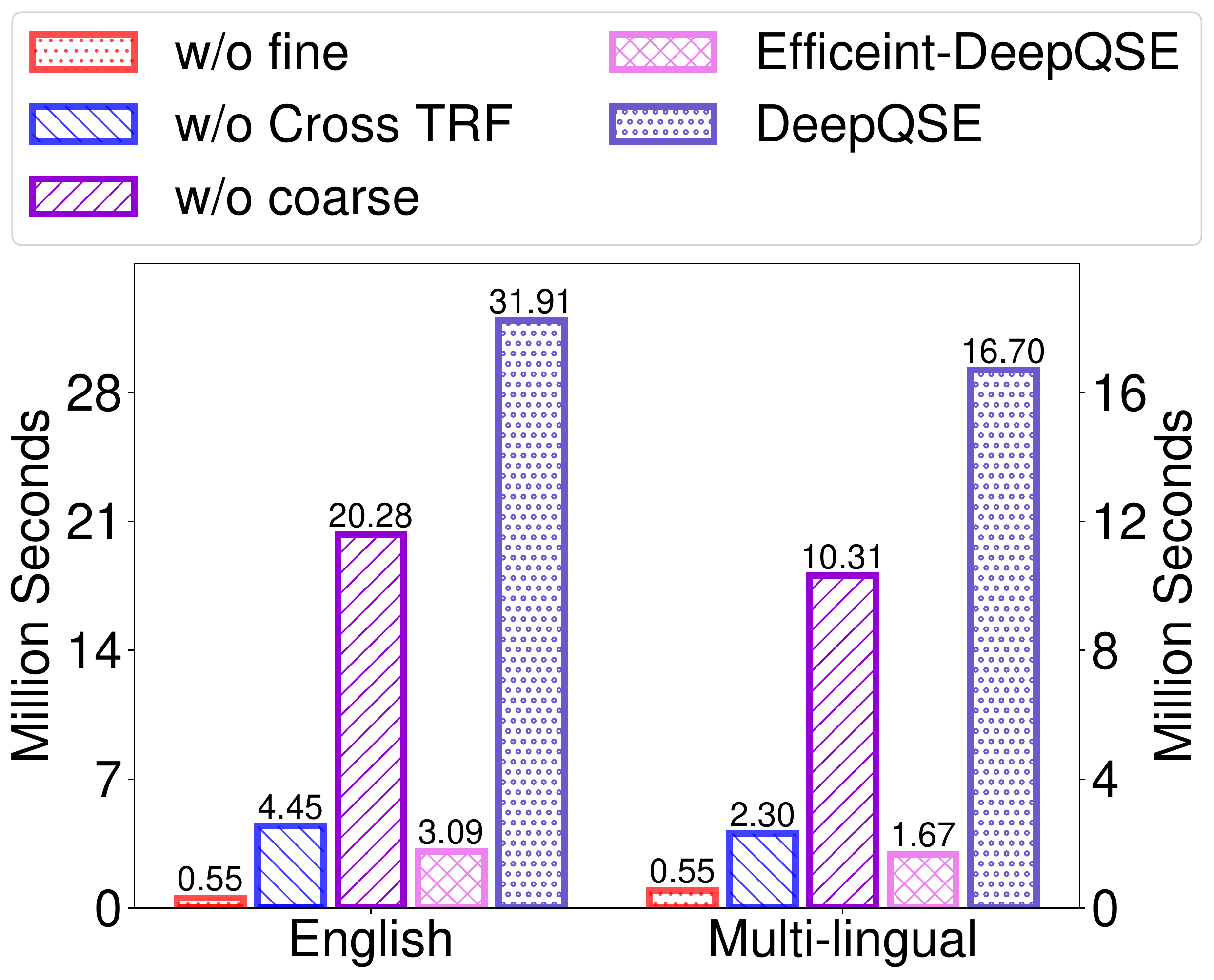}
  \caption{The impact of two-stage model and Cross Transformer on efficiency (ms).}
  \label{fig:two-step-efficiency-ms}
\end{figure}

In this subsection, we analyze how the Efficient-DeepQSE reduces the computation overhead of DeepQSE with a minor performance drop.
Compared with DeepQSE, the core improvement of Efficient-DeepQSE is the Cross Transformer, the coarse-grained candidate sentence selector and the fine-grained relevance model.
We remove these modules separately and show their performance and efficiency in Figure~\ref{fig:two-step-perform}, Figure~\ref{fig:two-step-efficiency} and Figure~\ref{fig:two-step-efficiency-ms}.
We have several observations from the results.
First, Efficient-DeepQSE has lower performance and lower computation overhead without the fine-grained relevance model.
This is because the fine-grained relevance model captures the deep semantic relevance between queries, titles and sentences, which can improve the performance.
And without the fine-grained relevance model, Efficient-DeepQSE does not need to perform the second stage, which lowers the computation overhead.
Second, Efficient-DeepQSE can achieve higher performance but higher computation overhead without the coarse-grained candidate sentence selector.
This is because the coarse-grained candidate sentence selector may select candidate sentences incorrectly, which increases the error rate.
However, it helps decrease the input size of the second stage.
Therefore, the computation overhead gets higher without the coarse-grained candidate sentence selector.
Finally, the computation overhead is higher without Cross Transformer.
This is because in Cross Transformer we only compute the query and title representations once, which avoids the repetitive computation in DeepQSE.
Combined with these components, the Efficient-DeepQSE reduces the computation overhead and achieves comparable performance with DeepQSE.

\subsection{Ablation Study}
\label{ref:efficient}
\begin{figure}[!t]
  \centering
  \includegraphics[width=0.37\textwidth]{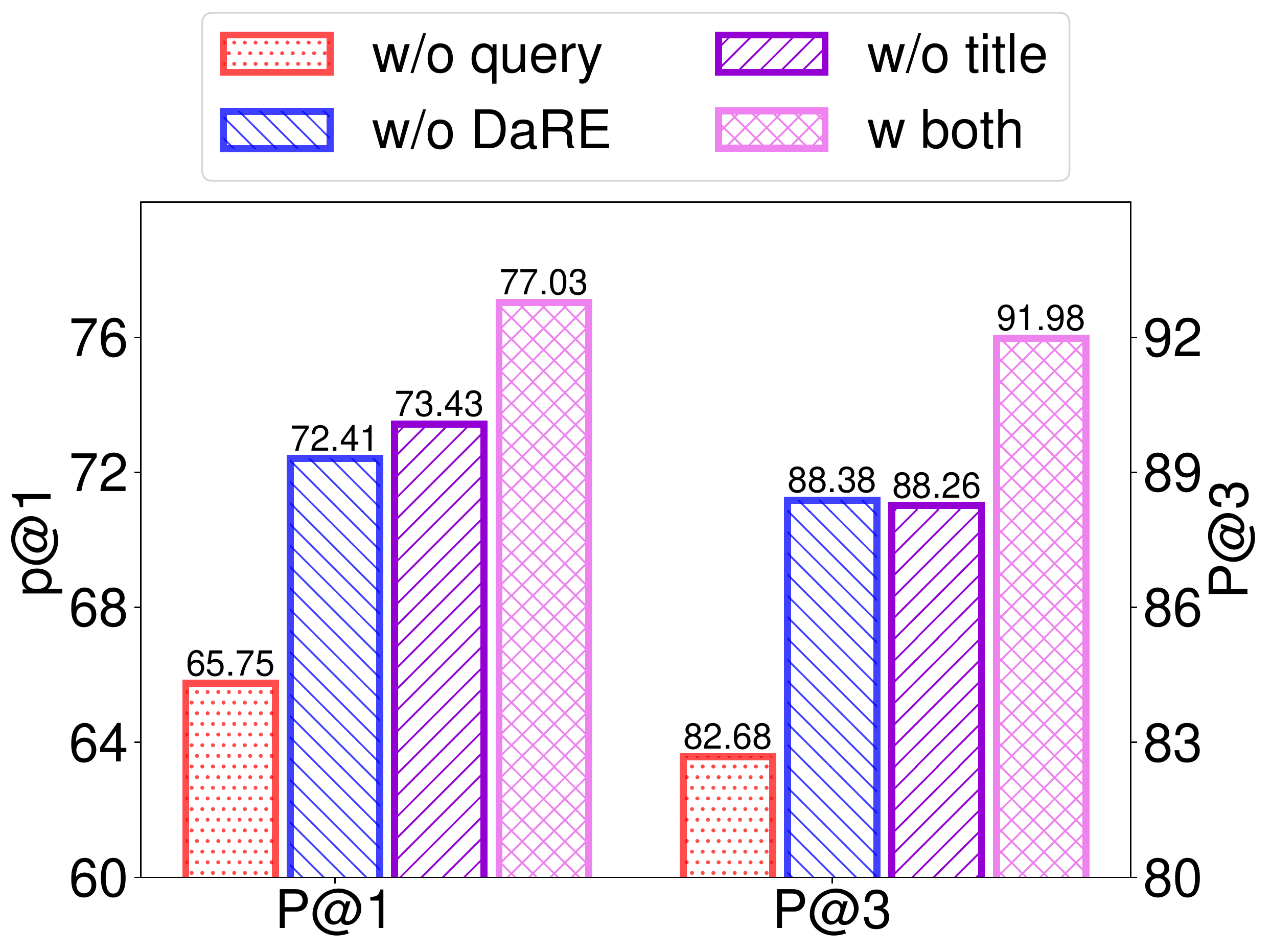}
  \caption{The impact of title, query and document-aware relevance encoder (DaRE) on \textit{English} dataset.}
  \label{fig:title}
\end{figure}

\begin{figure}[!t]
  \centering
  \includegraphics[width=0.37\textwidth]{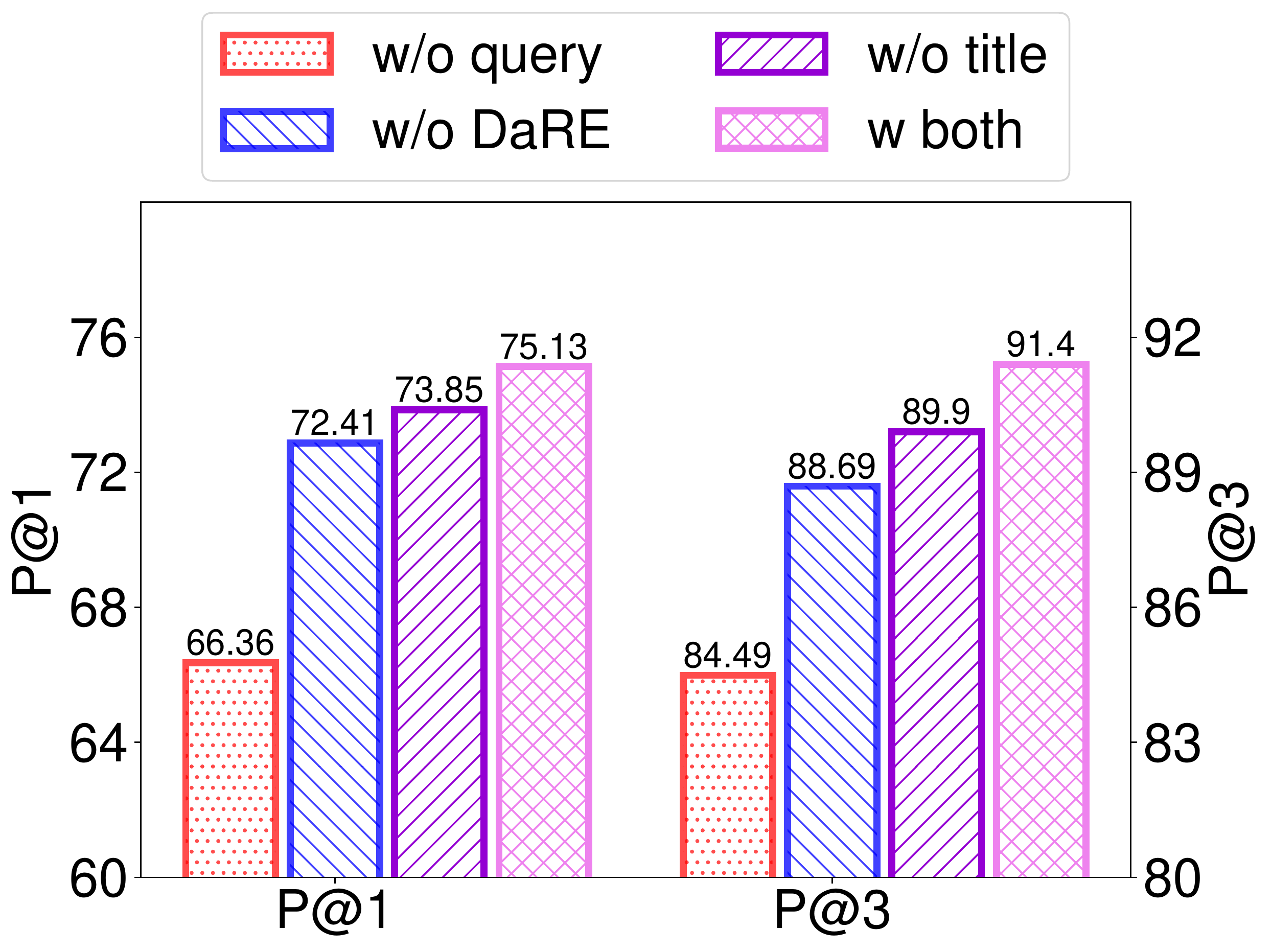}
  \caption{The impact of title, query and document-aware relevance encoder (DaRE) on \textit{Multi-lingual} dataset.}
  \label{fig:title-multi}
\end{figure}

In this section, we analyze the impact of adding titles, queries and the document-aware relevance encoder.
Due to space limitation, we only show the experimental result on Efficient-DeepQSE.
The same phenomenon can be observed on DeepQSE as well.
The results are shown in Figure~\ref{fig:title} and Figure~\ref{fig:title-multi}.
From the results, we can observe that the performance of Efficient-DeepQSE gets lower without titles.
This is because the titles can be treated as brief abstracts of webpages, which can help the model select sentences better summarizing the webpage content~\cite{10.1145/1321440.1321518}. 
It is also observed that the performance drops without queries.
This is because the selected snippets should not only summarize the webpage content, but also be relevant to queries.
Finally, the document-aware relevance encoder (DaRE) benefits snippet extraction.
This may be because the document-aware relevance encoder can model the query-document relevance in the context of the whole webpage, which helps select snippets better summarizing the webpage content.
\section{Conclusion}

In this paper, we propose a query-aware snippet extraction model for web search named DeepQSE.
DeepQSE first learns a query-aware sentence representation by modeling fine-grained interactions between queries, titles and sentences, then learns document-aware sentence relevance representations for snippet extraction.
To lower the computation overhead of DeepQSE, we further design the Efficient-DeepQSE, where the snippet extraction is decomposed into two stages, i.e. coarse-grained candidate sentence selection and fine-grained relevance modeling.
The coarse-grained selector can cache the sentence representations for fast online serving and parse several candidate sentences to the fine-grained relevance model.
In the fine-grained relevance model, we further design a Cross Transformer, to avoid the repetitive computation of query and title representations for different sentences.
Extensive experiments validate the effectiveness and efficiency of our approach.
\section{Limitations}
Our DeepQSE is a cross-encoder-based snippet extraction method.
It has great performance but heavy computation overhead, which is not beneficial for online inference.
We further propose Efficient-DeepQSE, an efficient version of DeepQSE, which divides the snippet extraction into two stages.
Although the Efficient-DeepQSE keeps the performance of DeepQSE and has much lower computation overhead than other PLM-based methods, it still has larger computation overhead than the conventional shallow-model-based methods.
We plan to further improve the efficiency of the snippet extraction algorithm in the future.
% \section*{Acknowledgments}

% % TODO
% We would like to thank Hao Wang for his suggestions on experiments.

\bibliography{anthology,custom}
\bibliographystyle{acl_natbib}

\appendix

\clearpage
\newpage

\noindent\begin{minipage}{\textwidth}
\captionsetup{type=figure}
\centering
\includegraphics[width=0.90\textwidth]{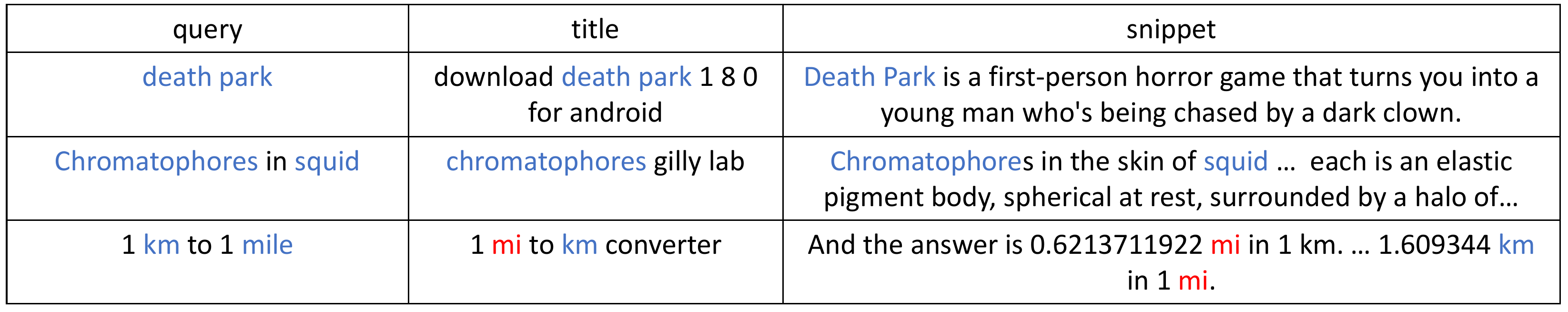}
\caption{Some snippets extracted by DeepQSE.}
\label{fig:case-qse}
\end{minipage}
% \vspace*{10\baselineskip}
\noindent\begin{minipage}{\textwidth}
\captionsetup{type=figure}
\centering
\includegraphics[width=0.90\textwidth]{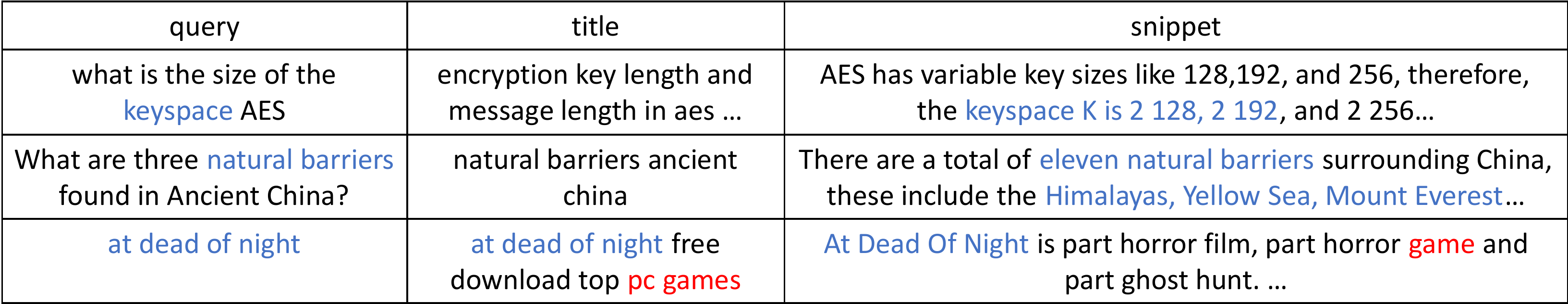}
\caption{Some snippets extracted by Efficient-DeepQSE.}
\label{fig:case}
\end{minipage}
% \vspace*{2.5\baselineskip}

% \begin{figure*}[h]
%   \vspace*{2.5\baselineskip}
%   \centering
%   \includegraphics[width=0.90\textwidth]{figure/case-qse.pdf}
%   \caption{Some snippets extracted by DeepQSE.}
%   \label{fig:case-qse}
% \end{figure*}
% \begin{figure*}[h]
%   \centering
%   \includegraphics[width=0.90\textwidth]{figure/case.pdf}
%   \caption{Some snippets extracted by Efficient-DeepQSE.}
%   \label{fig:case}
% \end{figure*}

\section*{Appendix}
\label{sec:appendix}
\subsection*{Detailed Data Construction Steps}
\label{sec:appendix-data}
The detailed data construction steps are as follows:
\textbf{Collect manually labeled dataset:} Given a pair of candidate snippets extracted from a document, human evaluators are asked to select the more appropriate one according to the corresponding query.
The manually labeled dataset can be formulated as $\{(q_i, s_{i1}, s_{i2}, d_i, l_i)|0\le i < M\}$, where $M$ is the number of samples in the dataset, $q_i$ is the query, $s_{i1}$ and $s_{i2}$ are candidate snippets, $d_i$ is the document, $l_i \in \{0, 1\}$ is the label. $l_i$ equals $0$ when $s_{i1}$ is more suitable than $s_{i2}$, otherwise $l_i$ equals $1$.
At least three annotators are assigned for a sample.

\textbf{Build snippet extraction dataset:} Since the manually labeled dataset is for pair-wise selection and is small for large PLM-based models, we train an ensemble model with the dataset to extract snippets for different documents according to corresponding queries. The extracted samples with high confidence scores are then used as the snippet extraction dataset.

\subsection*{Impact of Candidate Number}
\begin{figure}[!t]
    \vspace*{16\baselineskip}
    \centering
    \subfigure[\textit{English}]{\label{fig:k}\includegraphics[width=0.23\textwidth]{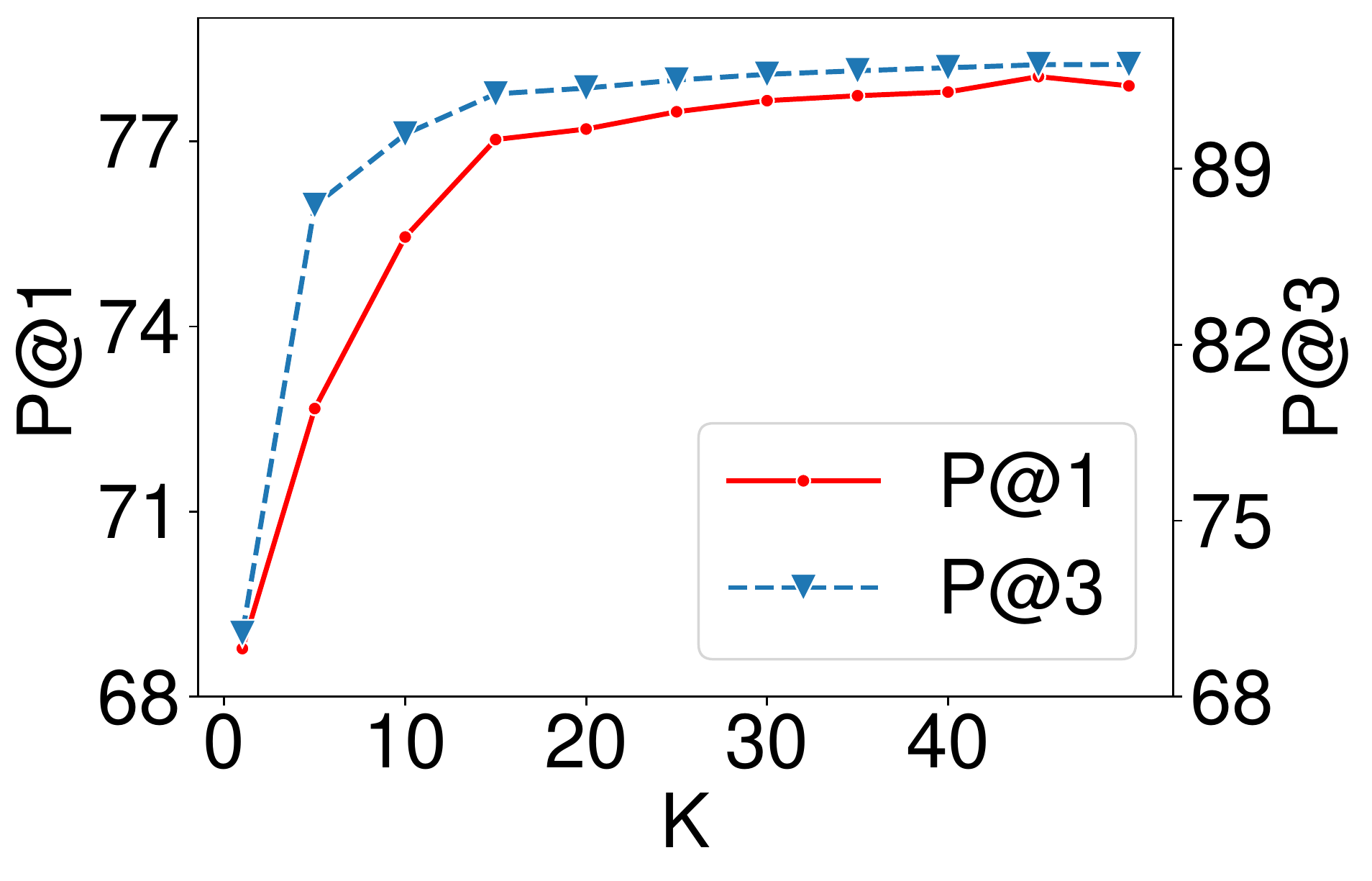}}
    \subfigure[\textit{Multi-Lingual}]{\label{fig:k-multi}\includegraphics[width=0.23\textwidth]{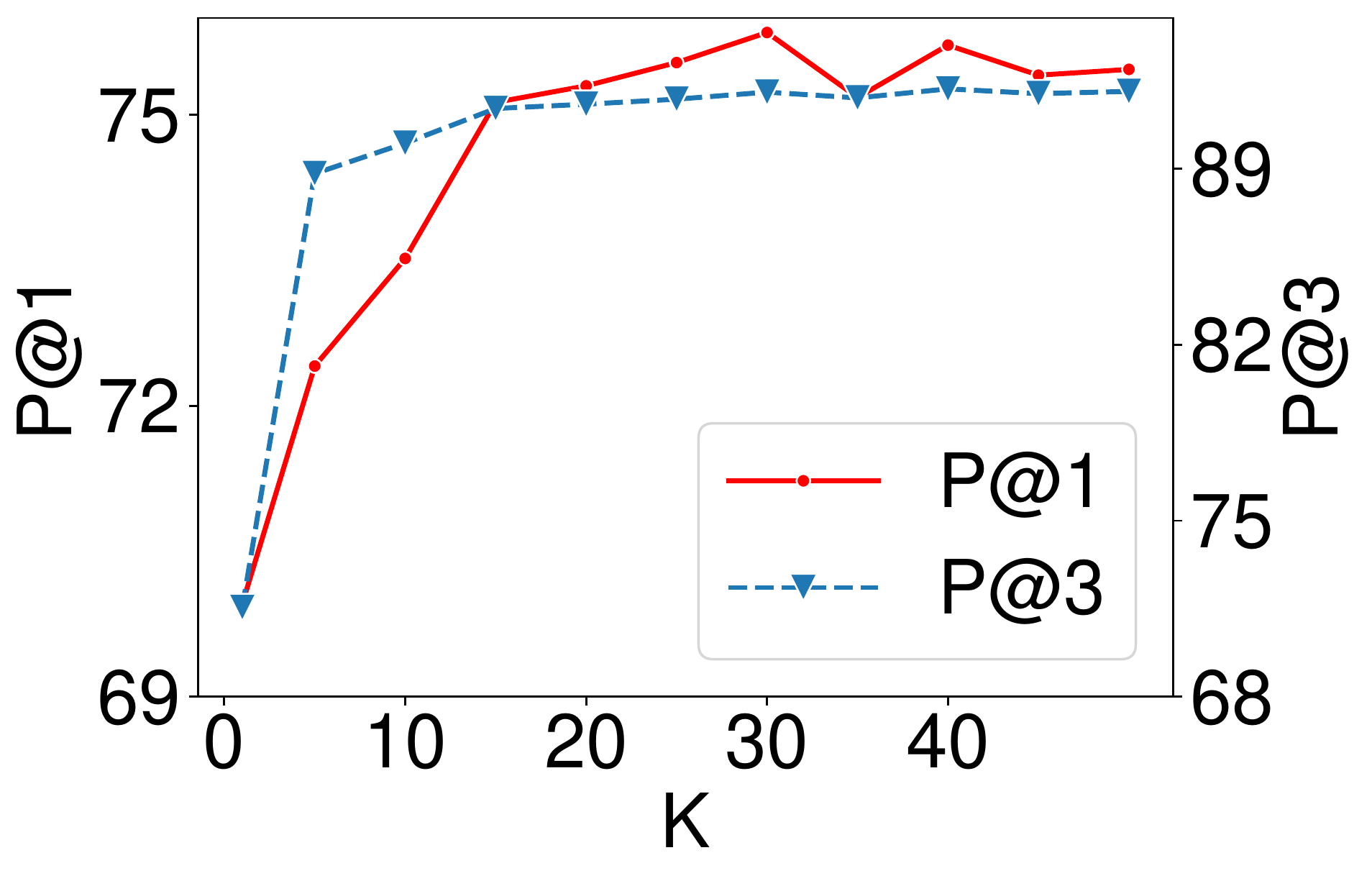}}
    \caption{Impact of candidate number.}
\end{figure}

In this subsection, we study the impact of the number of candidate sentences selected by the coarse-grained selector.
The experimental results are shown in Figure~\ref{fig:k} and Figure~\ref{fig:k-multi}.
We observe that with larger candidate sentence number, the performance of Efficient-DeepQSE is higher.
This may be because the probability that the ground truth sentence is selected in the candidate sentences gets higher.
However, the computation overhead of Efficient-DeepQSE linearly increases with 
larger candidate sentence number.
How to choose a proper candidate sentence number to achieve a great trade-off between performance and efficiency is the key point of our method.

% \begin{figure}[!t]
%   \vspace*{16\baselineskip}
%   \centering
%   \includegraphics[width=0.30\textwidth]{figure/K-performance.pdf}
%   \caption{Impact of candidate number on \textit{English} dataset.}
%   \label{fig:k}
% \end{figure}
% \begin{figure}[!t]
%   \centering
%   \includegraphics[width=0.30\textwidth]{figure/K-performance-multi.pdf}
%   \caption{Impact of candidate number on \textit{Multi-Lingual} dataset.}
%   \label{fig:k-multi}
% \end{figure}

\subsection*{Case Study}
In this subsection, we show some snippets extracted by our DeepQSE in Figure~\ref{fig:case-qse} and our Efficient-DeepQSE in Figure~\ref{fig:case}.
In all cases, the snippets are relevant to the input query.
This is because we model the word-level interactions between query and sentences in DeepQSE and Efficient-DeepQSE.
Meanwhile, the selected snippets summarize the webpage content in the context of the input query.
This is because we consider the context of webpage in the document-aware relevance encoder, which enables our method to capture the global webpage information.

\subsection*{Experimental Environments}
We conduct experiments with a Linux server with 8 V100 GPUs with 32GB memory.
The version of CUDA is 11.1.
We implement both DeepQSE and Efficient-DeepQSE with pytorch 1.9.1.

\end{document}